\documentclass[10pt,twocolumn,letterpaper]{article}

\usepackage[pagenumbers]{cvpr} 

\definecolor{cvprblue}{rgb}{0.21,0.49,0.74}
\usepackage[pagebackref,breaklinks,colorlinks,allcolors=cvprblue]{hyperref}

\usepackage{hyperref}
\hypersetup{
	colorlinks=true,
	citecolor=teal,
}
\usepackage{url}

\usepackage{amsmath}
\usepackage{booktabs} 
\usepackage{multirow} 
\usepackage{multicol} 
\usepackage{graphicx}
\usepackage{booktabs}

\usepackage{pifont}
\usepackage{enumitem}
\usepackage{wrapfig}
\usepackage{colortbl}
\definecolor{usercolor}{RGB}{184,85,10}
\definecolor{assistantcolor}{RGB}{0,115,0}  

\let\oldding\ding
\renewcommand{\ding}[2][1]{\scalebox{#1}{\oldding{#2}}}

\def\logo{\makebox[0pt][l]{\hspace{0pt}\raisebox{-0.5ex}{\includegraphics[scale=0.03]{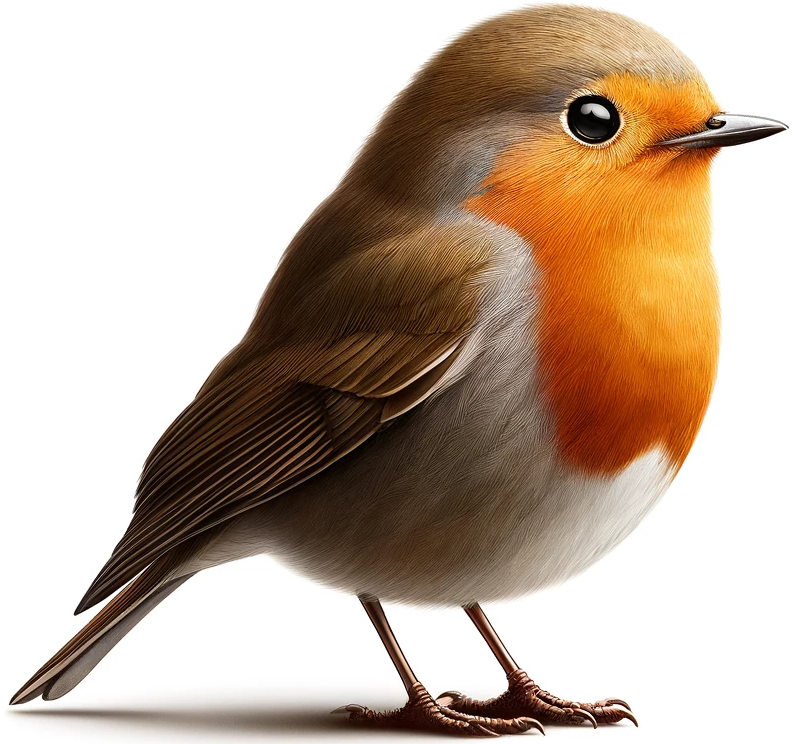}}}}

\title{Robin3D\logo~~~~~~:~Improving 3D Large Language Model \\
via Robust Instruction Tuning}

\author{Weitai Kang$^{1}$\thanks{Code: \url{https://github.com/WeitaiKang/Robin3D}}, Haifeng Huang$^{2}$, Yuzhang Shang$^{4}$, Mubarak Shah$^{3}$, Yan Yan$^{1}$ \\
$^1$University of Illinois Chicago, $^2$Zhejiang University, \\
$^3$University of Central Florida, $^{4}$Illinois Institute of Technology
}

\begin{document}
\maketitle
\begin{abstract}
Recent advancements in 3D Large Language Models (3DLLMs) show their potential to build general-purpose agents in the 3D real world,
yet challenges remain due to the lack of high-quality robust instruction-following data, leading to limited discriminative power and generalization of 3DLLMs. In this paper, we introduce Robin3D, a powerful 3DLLM trained on large-scale instruction-following data generated by our novel data engine, Robust Instruction Generation (RIG) engine. RIG generates two key instruction data: 1) the Adversarial Instruction-following data, which features mixed negative and positive samples to enhance the model's discriminative understanding. 2) the Diverse Instruction-following data, which contains various instruction styles to enhance model's generalization. 
As a result, we construct \textbf{1 million} instruction-following data, consisting of 344K Adversarial samples, 508K Diverse samples, and 165K benchmark training set samples.
To better handle these complex instructions, 
Robin3D further integrates an improved vision projector and enhanced sequence organization.
Notably, we achieve a \textbf{7.8\%} improvement in the grounding task (Multi3DRefer) and a \textbf{6.9\%} improvement in the captioning task (Scan2Cap).
\end{abstract}

\section{Introduction}
Spatial Intelligence \citep{Li2024TED} refers to the ability of AI to understand the 3D world and reason within 3D space. Related ideas, such as Embodied AI \citep{embodiedai} and Robotic Agent \citep{robocat}, express a similar aim to build general-purpose assistants in the 3D real world. 
To achieve this goal, researchers have drawn inspiration from the success of 2D Multimodal Large Language Models (MLLMs) \citep{llava, ferret} and have started exploring the potential of 3D Large Language Models (3DLLMs)~\citep{3dllm, ll3da, chat3d, chat3dv2}.

\begin{figure}
    \centering
    \includegraphics[width=0.46\textwidth]{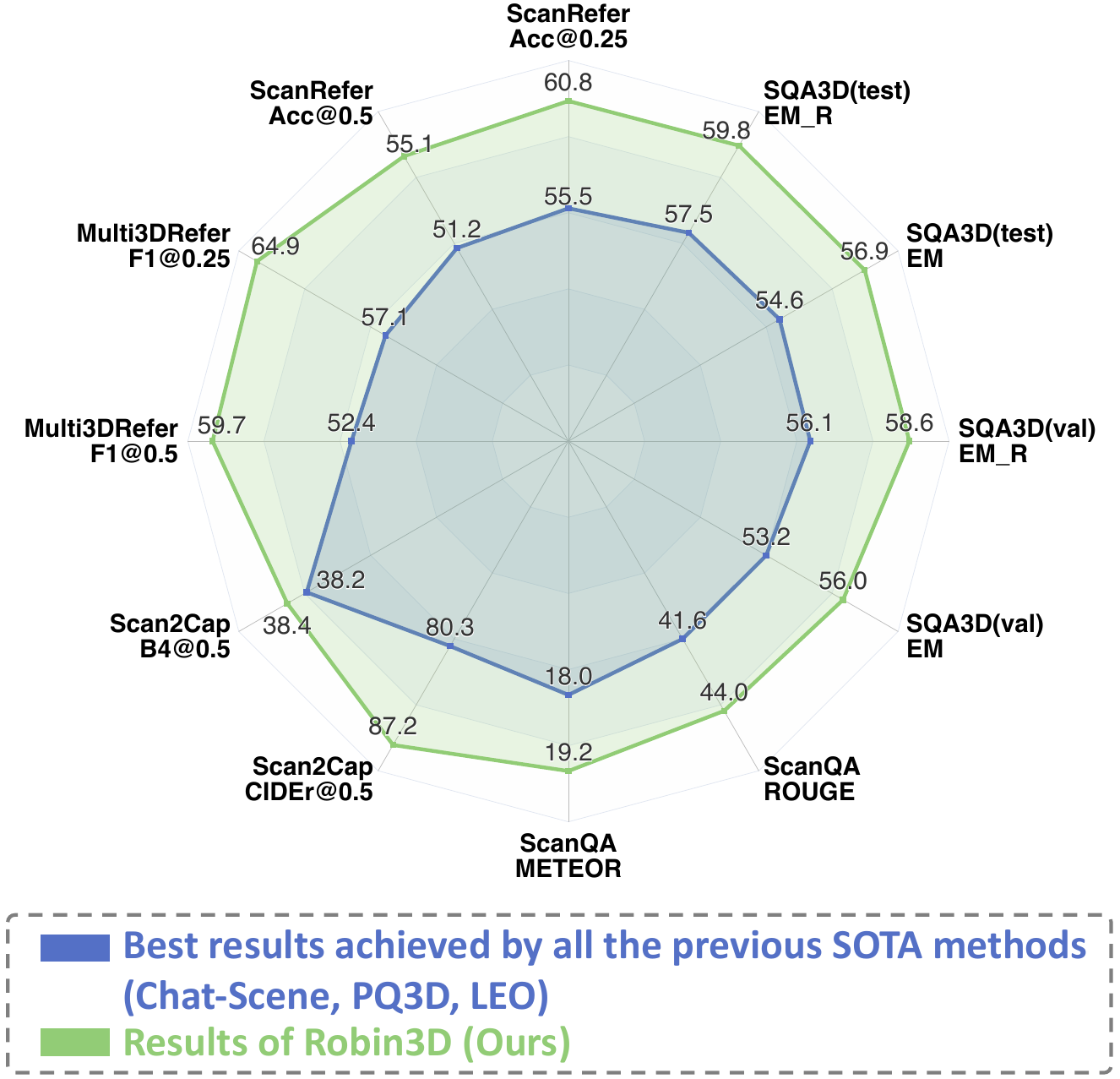}
    \vspace{-8pt}
    \caption{Robin3D surpasses previous SOTA on all 12 benchmarks by training on our RIG-generated 1 million data.}
    \label{radar_data}
    \vspace{-15pt}
\end{figure}

Instruction-following tuning \citep{llava, llava1.5, llavanext} in MLLMs refers to training the LLM to execute natural language commands by integrating both textual and visual information. In contrast to the versatile image-text pairs employed for training 2D MLLMs, collecting 3D instruction-following data for 3DLLM remains a significant challenge. Although existing works have made progress \citep{3dllm,leo,grounded3dllm} in generating more instruction data, they still lack robustness in two aspects: {\bf 1)} \textit{Most of the existing instruction data consist of positive pairs, lacking adversarial or negative samples.} Therefore, a concern remains that models trained on such data tend to be less discriminative, as they risk overfitting to the positive pairs and are more likely to hallucinate positive responses, regardless of the input. {\bf 2)} \textit{Current instruction data lack diversity in language styles}, as human annotators or generative models \citep{chatgpt, CogVLM} are typically asked to follow fixed instructions when describing objects \citep{scanrefer, scan2cap}, or the data is generated using predefined templates \citep{nr3d}, which may limit models' generalizability.

To address these challenges, we introduce Robin3D, a robust and powerful 3D Large Language Model tuned on large scale instruction-following data generated by our novel data engine, Robust Instruction Generation (RIG) engine. Specifically, RIG is designed to generate two types of data: 
\raisebox{-1.1pt}{\ding[1.1]{182\relax}} 
Adversarial Instruction-following data, which is characterized by mixing adversarial or negative samples with positive ones. This process decouples the potential memorized positive pairs in the training set, leading to a more discriminative understanding of individual objects and instructions.
To present a comprehensive adversarial dataset, we cover both object-level and scene-level instructions, from category-based identification problems to expression-based reasoning challenges, resulting in four new tasks.
\raisebox{-1.1pt}{\ding[1.1]{183\relax}} Diverse Instruction-following data, which first comprehensively collects various types of instructions from existing studies or transforms current tasks into instruction-following format.
To harness the powerful in-context learning capability of LLMs, we use ChatGPT~\citep{chatgpt} to diversify the language styles of the instructions by crafting specific prompt engineering tailored to each task.
Combining these with the original training sets of current benchmarks, we construct \textbf{1 million} instruction-following samples, with approximately 344K adversarial data, 508K diverse data, and 165K benchmark data.

We choose Chat-Scene \citep{chat-scene} as the bedrock of Robin3D's structure due to its efficient extraction of 2D-3D features and its unified approach for object referring and grounding via object IDs. 
To better handle challenging data from RIG, which contains complex referring and grounding requirements, we incorporate an enhanced vision projector and a more informative way to organize the sequence of tokens. Specifically, a Relation-Augmented Projector is proposed to enrich object-centric features with scene-level context. Additionally, we strengthen the connection between object IDs and object features by wrapping features with identical ID tokens and reinforcing this link via a post-vision order.

In sum, we introduce Robin3D, a powerful 3DLLM trained on robust instruction-following data generated by our novel data engine, RIG. To better handle our generated data, we further involve improvements on the vision projector and sequence organization. 
As shown in \cref{radar_data}, Robin3D surpasses previous SOTA on all benchmarks without the need for task-specific fine-tuning, different from previous approaches \citep{3dllm, ll3da}. 

\section{Related Work}
\paragraph{3D Vision-Language Learning}
Recent advancements in 3D vision-language (3D-VL) learning \citep{scanrefer, nr3d, scanqa, scan2cap, intent3d} have focused on bridging the gap between 3D scene understanding and natural language. Similar to 2D domain \citep{referitgame, actress, densecap, ferret, segvg, vqa, ferretv2, attbalance}, tasks like 3D Visual Grounding \citep{scanrefer, multi3drefer, nr3d}, 3D Dense Captioning \citep{scan2cap}, and 3D Question Answering \citep{scanqa, sqa3d} have been proposed to evaluate models’ ability to understand human instructions related to 3D objects. Early methods focus on single-task models, such as EDA \citep{eda} for grounding and Vote2Cap-DETR++ \citep{vote2cap} for captioning. 
Some studies also develop unified models capable of handling multiple 3D scene-language tasks. Approaches like 3DJCG \citep{3djcg} and D3Net \citep{d3net} leverage task synergies, while 3D-VisTA \citep{3dvista}, 3D-VLP \citep{3dvlp} and PQ3D \citep{pq3d} introduce pre-training techniques and unified representations to align 3D vision features with language. However, their dependence on task-specific heads restricts their flexibility for more generalized user-assistant interactions.

\begin{figure*}
    \centering
    \includegraphics[width=0.9\textwidth]{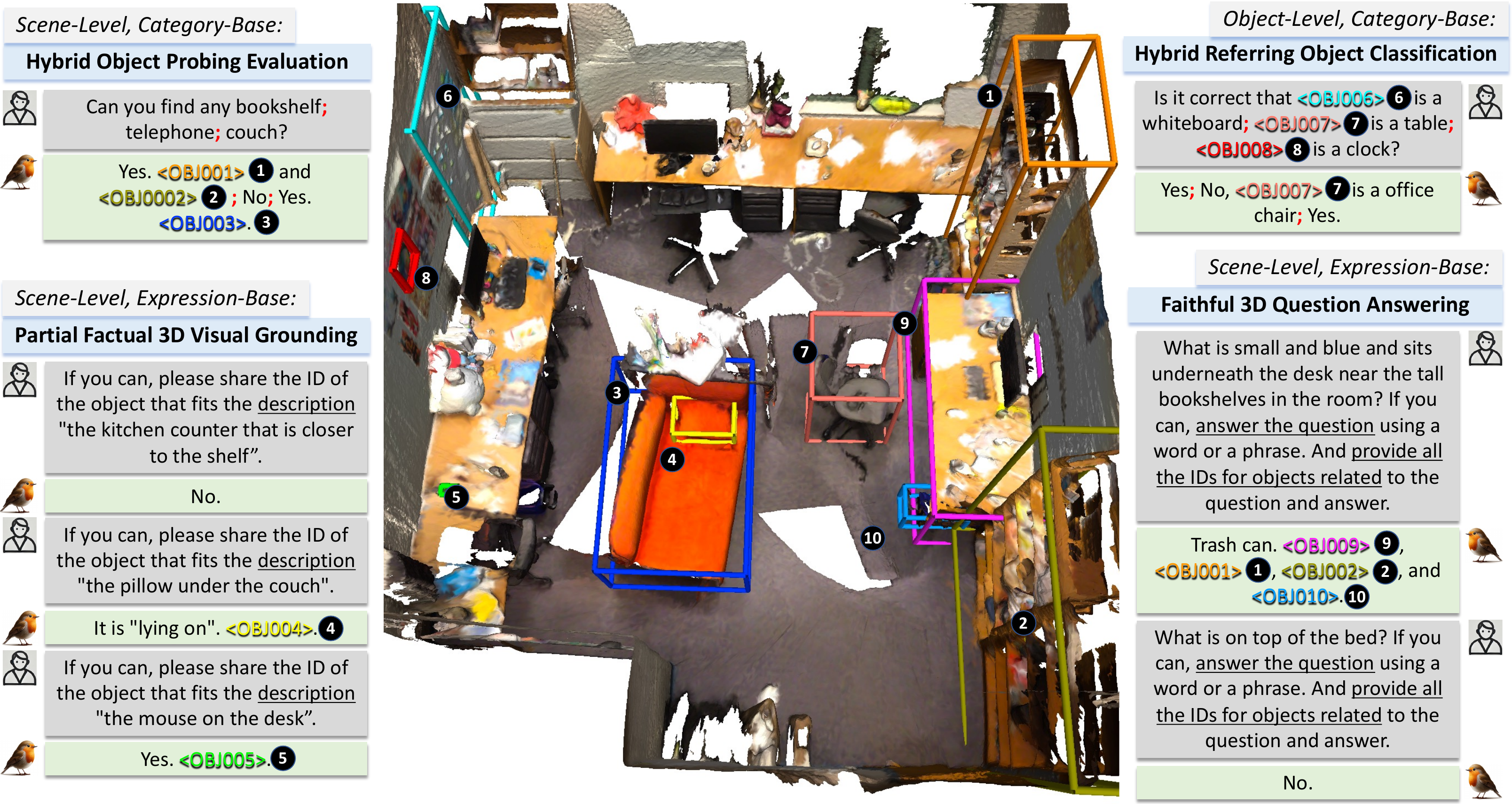}
    \vspace{-10pt}
    \caption{The visualization of examples of adversarial / negative data. For better visualization, we associate each object ID with the same color as its bounding box. The black solid circles with numbers are solely for visualization purposes and are not included in the actual data.}
    \label{adversarial}
    \vspace{-10pt}
\end{figure*}

\paragraph{3D Large Language Model}
Following the success of 2D MLLM \citep{ferret, ferretv2, llava, llava1.5, llavanext}, researchers begin to explore MLLM in 3D domain (3DLLM).
Models like PointLLM \citep{pointllm} and Imagebind-LLM \citep{imagebind} show strong performance in object-level tasks by mapping 3D data into LLMs. However, they face difficulties in handling scene-level reasoning. 
3D-LLM \citep{3dllm} incorporates positional embeddings and location tokens,
and Oryx \citep{oryx} offers a solution to support multi-view arbitrary resolution images.
However, their reliance on 2D encoders limits the ability to fully capture 3D spatial structures. Models such as LL3DA \citep{ll3da}, Chat-3D \citep{chat3d}, LEO \citep{leo}, and Scene-LLM \citep{scenellm} have made progress in improving scene-level dialogue capabilities, showing promising results in question-answering and captioning tasks. 
However, their insufficient visual grounding capability limits their application in Embodied AI or Robotic Agents, which require precise object localization and manipulation following human instruction.
To further enhance grounding abilities, Grounded 3D-LLM \citep{grounded3dllm} introduces referent tokens and the contrastive learning to unify grounding and textual responses. Similarly, Chat-3D v2 \citep{chat3dv2} proposes the use of object identifiers (object IDs) for referring and grounding. Building on Chat-3D v2\citep{chat3dv2}, Chat-Scene \citep{chat-scene} further incorporates DINO v2 \citep{dinov2} to provide strong multi-view, object-centric 2D representations.
Despite these advancements, current 3D LLMs, which are trained solely on positive 3D vision-language pairs and template-based instructions, suffer from suboptimal generalization and a potential for overfitting.
\section{Approach}

\subsection{Preliminary}
As a strong baseline for our Robin3D structure, Chat-Scene \citep{chat-scene} demonstrates commendable proficiency across multiple benchmarks, attributed to its efficient feature extraction and the use of object IDs for object referring and grounding. They use the pre-trained Mask3D \citep{mask3d} to predict 3D masks for each object. Based on these masks, the point clouds for each object are sampled, normalized, and processed by the pre-trained Uni3D \citep{uni3d} to extract unified, object-centric 3D features. Additionally, 2D masks projected from the 3D masks are used to sample and average 2D features, which are extracted by the pre-trained DINO v2 \citep{oquab2023dinov2} from multi-view images of each object. Special tokens ${<\texttt{OBJ}i>}_{i=1...n}$ are added to the vocabulary as object IDs and interleaved with the above 3D and 2D object features in the input sequence of LLM to indicate each object for referring in the input and grounding in the output. 

However, Chat-Scene only trains benchmark data that are all positive pairs and are labeled or generated by fixed formats, leading to be less discriminative. We detail our generation of robust instruction data in the following and then introduce our improvements on the vision projector and sequence organization to better handle our robust data.

\subsection{Robust Instruction Generation (RIG)}

\subsubsection{Adversarial Data Generation}
The Adversarial data is designed to challenge the model's discriminative capabilities by introducing adversarial or negative samples, ranging from the object-level to the scene-level. It features both category-based identification tasks and expression-based reasoning challenges. As shown in Fig.~\ref{adversarial}, we ultimately formulate four novel tasks: Hybrid Object Probing Evaluation, Hybrid Referring Object Classification, Partial Factual 3D Visual Grounding, and Faithful 3D Question Answering.
Details are as follows:

\paragraph{Hybrid Object Probing Evaluation (HOPE) -- \textit{Fig.~\ref{adversarial}(upper left)}}\label{HOPE}
To construct a scene-level category-based task, we introduce HOPE, which is inspired by the POPE benchmark \citep{pope} in 2D domain. POPE evaluates the tendency of 2D MLLMs to hallucinate by asking yes/no questions about the presence of one specific object at a time. Building on this, HOPE further incorporates such hallucination challenges into the training stage in the 3D domain, aiming to train our model to be more discriminative. Additionally, HOPE presents a hybrid scenario, introducing greater complexity to further advance the decoupling of memorized positive vision and language pairs. Specifically, given a 3D scene, we ask the model to determine the presence of various randomly specified objects. The objects may or may not be present in the scene, and each existing object might have one or more instances. The model is required to answer ``{\em No}'' when the object is not present in the scene, and answer ``{\em Yes}'' with the object ID of each instance of the object when it exists. 
As shown in Fig.~\ref{adversarial} (upper left), the question combines multiple objects, separated by semicolons (``;''), and the answer combines responses for each object, also separated by semicolons. This structure creates a challenging setting that involves hybrid recognition of both positive and negative object presence, combined with multi-instance object localization.

\paragraph{Hybrid Referring Object Classification (HROC) -- \textit{Fig.~\ref{adversarial}(upper right)}}\label{HROC}
 
Referring Object Classification \citep{ferret} evaluates a model's ability to understand a referred region in 2D domain, 
focusing on a classification problem by ``Region-in Text-out'' format. Our HROC dataset extends this task into the training data for 3D domain to create an object-level category-based task,
by incorporating adversarial and hybrid challenges. 
Specifically, in a 3D scene, we randomly create hybrid positive and negative ID-Category pairs to form our questions, as illustrated in Fig.~\ref{adversarial} (upper right). A positive pair consists of a valid object ID and the ground truth category. The bounding box of the corresponding object ID must overlap with one ground truth bounding box, and the category of the overlapping object is defined as the ground truth category. A negative pair includes a valid object ID and a randomly selected category that is present in the scene but not the ground truth category to serve as an adversarial challenge. The model is required to answer ``{\em Yes}'' for positive pairs and ``{\em No}'' with the correct category for negative pairs. The pairs and corresponding answers are separated by semicolons (``;'').

\paragraph{Partial Factual 3D Visual Grounding (PF-3DVG) -- \textit{Fig.~\ref{adversarial}(lower left)}}\label{PF-3DVG}

Our PF-3DVG introduces a scene-level expression-based task, featuring three types of data in 3DVG: unfactual data, partially factual data, and factual data. For unfactual data, given a 3D scene, we randomly select a reference from Sr3D+ \citep{nr3d} where the indicated object does not exist in the scene. The model is required to answer ``{\em No}'' when prompted with the question, as shown in the first example of Fig.~\ref{adversarial} (lower left).
For partial factual data, given a reference from Sr3D+ and its corresponding 3D scene, we randomly switch the described spatial relationship with a different one based on the predefined template of Sr3D+. For example, as shown in the second example of Fig.~\ref{adversarial} (lower left), we change the original reference ``{\em the pillow lying on the couch}'' to ``{\em the pillow under the couch}''. In this case, it is still possible for human annotators to ground the target based on this partial factual information, but this introduces an adversarial challenge since the information is not completely accurate.
Therefore, we require the model to retify the information and answer ``{\em It is `lying on'}'' while providing the grounding result (object ID). Notably, we only use references whose target object has no distractors sharing the same category, ensuring that the partial factual information is still informative enough for grounding the target and does not lead to ambiguity.
For factual data, we randomly augment the original spatial relationship with its synonym to improve diversity. For example, the synonym of ``{\em below}'' can be ``{\em under}'', ``{\em beneath}'', or ``{\em underneath}''.

\begin{figure}[t]
    \centering
    \includegraphics[width=0.48\textwidth]{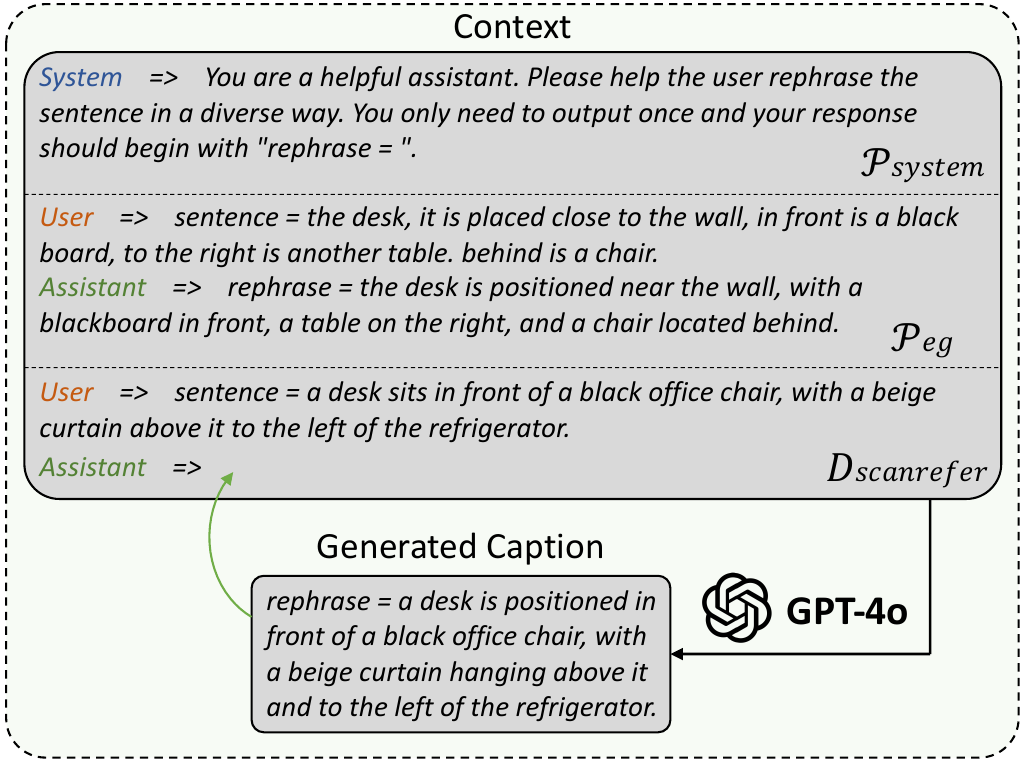}
    \caption{Pipeline to generate our Diverse Instruction data by the in-context learning of ChatGPT.}
    \label{prompt1}
\end{figure}
\begin{figure}[t]
    \centering
    \includegraphics[width=0.48\textwidth]{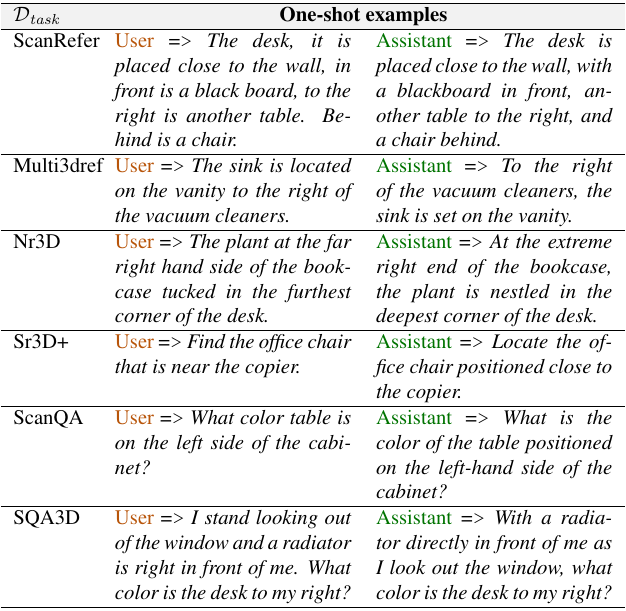}
    \caption{The one-shot examples for ChatGPT to rephrase the instruction-following data.}
    \label{prompt2}
\end{figure}

\paragraph{Faithful 3D Question Answering (3DFQA) -- \textit{Fig.~\ref{adversarial}(lower right)}}\label{3DFQA}
The original 3D Question Answering (QA) task \citep{scanqa} includes only positive samples, which can potentially lead to the model memorizing fixed combinations of 3D scenes and QA pairs. To address this, we propose Faithful 3D Question Answering, a scene-level expression-based task which incorporates both negative and positive examples with an additional grounding requirement.
To construct negative samples, we first sample a QA pair and collect the related objects that are mentioned in the question or the target objects of the answer from \citep{scanqa}. Then, we randomly select a 3D scene that lacks those related objects. A new instruction is added to the question: ``{\em If you can, answer the question... and provide all the IDs...}'' as illustrated in Fig.~\ref{adversarial} (lower right). In this case, the model must faithfully answer ``{\em No}'' based on the absence of related objects in the 3D scene and must not provide any object IDs, demonstrating its reliance on the scene for making decisions.
For positive samples, directly taken from \citep{scanqa}, the model must answer the question while faithfully grounding its ``evidence'' for the answer, i.e., providing the IDs of the related objects.
Therefore, the model trained on our 3DFQA dataset
is forced to generalize beyond memorization,
learning to respond faithfully to both positive and negative samples.

\subsubsection{Diverse Data Generation}
The Diverse data aim to enhances the model’s generalization by first incorporating multiple different types of instruction-following data and then increasing the linguistic diversity of the instructions.

We first collect large scale data from different tasks outside the benchmark dataset.
Specifically, given a 3D scene, we collect question-answering pairs from the following tasks: 1) Category Question-Answering task from \citep{chat3dv2}, where the model is asked to answer the category of a specified object. 2) Nr3D Captioning task from \citep{chat3dv2}, where the model is asked to caption the spatial relationship of a specified object to its neighboor. The ground truth is constructed from Nr3D \citep{nr3d} dataset. 3) Appearance Captioning task from \citep{grounded3dllm}, where the model is asked to captioning the physical attributes or visual characteristics of a specified object. The ground truth is generated by CogVLM \citep{CogVLM}. 4) Region Captioning task from \citep{grounded3dllm}, where the model is asked to caption the region encircling a specified object. The ground truth is organized by ChatGPT \citep{chatgpt}. 5) End to end 3D Visual Grounding from Nr3D dataset \citep{nr3d}, where the model is not provided ground truth candidates, different from the original setting in Nr3D. 6) End to end 3D Visual Grounding from Sr3D+ dataset \citep{nr3d}, where the model is also not provided ground truth candidates, different from the original setting in Sr3D+.

To diversify the wording style,
we develop a scalable pipeline by harnessing ChatGPT’s \citep{chatgpt} in-context learning ability to rephrase the above data. This is achieved through a combination of one-shot examples and structured prompt engineering, as shown in Fig.~\ref{prompt1}.
Formally speaking, given a collected instruction-following dataset $\mathcal{D}_{\mathit{task}}$, where $\mathit{task} \in \{\text{ScanRefer}, \text{Multi3DRefer}, \text{Nr3D}, \text{Sr3D+}$, $\text{Nr3D\ Captioning}, \text{ScanQA}, \text{SQA3D}, \text{PF-3DVG}, \text{3DFQA} \}$, we construct a system prompt, $\mathcal{P}_{\mathit{system}}$, to indicate the rephrase requirement and structured output format to ChatGPT, a one-shot example prompt, $\mathcal{P}_{\mathit{eg}}$, to show a rephrased example and output format for ChatGPT to better understand the requirement, and randomly assign a temperature $\mathcal{T}$ from $[1.1, 1.2, 1.3]$ for ChatGPT to increase the randomness of the output diversity. Our rephrased output, $\mathcal{D}_{\mathit{rephrase}}$, is generated by $\mathcal{D}_{\mathit{rephrase}} = \mathcal{M}(\mathcal{P}_{\mathit{system}}, \mathcal{P}_{\mathit{eg}}, \mathcal{D}_{\mathit{task}}, \mathcal{T})$, where $\mathcal{M}$ is the GPT-4o version of ChatGPT. We provide the details of $\mathcal{P}_{\mathit{system}}$ and $\mathcal{P}_{\mathit{eg}}$ in the Fig.~\ref{prompt1} for data of ScanRefer as an example. With our ``{\em sentence=}'' and ``{\em rephrase=}'' structured prompt, GPT-4o can easily follow the requirement and we can conveniently collect the output by detecing the ``{\em rephrase=}'' keywords.
In Fig.~\ref{prompt2}, we provide details regarding the one-shot example for each task. Since Nr3D Captioning is constructed from Nr3D, PF-3DVG is from Sr3D+, and 3DFQA is from ScanQA, we do not provide additional examples for them.

\begin{table*}[t]
\centering
\resizebox{\linewidth}{!}{
\begin{tabular}{lcccccccccccc}
\toprule
\multirow{2}{*}{\textbf{Model}} & \multicolumn{2}{c}{ScanRefer} & \multicolumn{2}{c}{Multi3DRefer} & \multicolumn{2}{c}{Scan2Cap} & \multicolumn{2}{c}{ScanQA(val)} & \multicolumn{2}{c}{SQA3D(val)} & \multicolumn{2}{c}{SQA3D(test)} \\ 
\cmidrule(lr){2-3} \cmidrule(lr){4-5} \cmidrule(lr){6-7} \cmidrule(lr){8-9} \cmidrule(lr){10-11} \cmidrule(lr){12-13}
 & \scalebox{0.85}[1]{Acc@0.25} & \scalebox{0.85}[1]{Acc@0.5} & \scalebox{0.85}[1]{F1@0.25} & \scalebox{0.85}[1]{F1@0.5} & \scalebox{0.85}[1]{B-4@0.5} & \scalebox{0.85}[1]{C@0.5} & \scalebox{0.85}[1]{M} & \scalebox{0.85}[1]{R} & \scalebox{0.85}[1]{EM} & \scalebox{0.85}[1]{EM-R} & \scalebox{0.85}[1]{EM} & \scalebox{0.85}[1]{EM-R} \\ \midrule
\multicolumn{1}{l}{\textit{\text{Task-Specific Training}}}\\  \addlinespace[0.1cm]
ScanRefer & 37.3 & 24.3 & - & - & - & - & - & - & - & - & - & - \\
EDA &  53.8 & 41.7 & - & - & - & - & - & - & - & - & - & - \\
Concretenet &  50.6 & 46.5 & - & - & - & - & - & - & - & - & - & - \\
M3DRef-CLIP & 51.9 & 44.7 & 42.8 & 38.4 & - & - & - & - & - & - & - & - \\
Scan2Cap & - & - & - & - & 23.3 & 39.1 & - & - & - & - & - & - \\
Vote2Cap-DETR++ & - & - & - & - & 37.1 & 67.6 & - & - & - & - & - & - \\
ScanQA & - & - & - & - & - & - & 13.1 & 33.3 & - & - & - & - \\
SQA3D & - & - & - & - & - & - & - & - & - & - & 46.6 & - \\ \midrule
\multicolumn{1}{l}{\textit{\text{Joint Training}}}\\  \addlinespace[0.1cm]
D3Net & - & 37.9 & - & 32.2 & 35.7 & 62.6 & - & - & - & - & - & - \\
3DJCG & 49.6 & 37.3 & - & 26.6 & 31.0 & 49.5 & - & - & - & - & - & - \\
3D-VLP & 51.4 & 39.5 & - & - & 32.3 & 54.9 & - & - & - & - & - & - \\
3D-VisTA & 50.6 & 45.8 & - & - & 34.0 & 66.9 & 13.9 & 35.7 & - & - & 48.5 & - \\
PQ3D & - & \underline{51.2} & - & 50.1 & 36.0 & \underline{80.3} & - & - & - & - & 47.1 & - \\
3DLLM & 30.3 & - & - & - & - & - & 14.5 & 35.7 & - & - & - & - \\
Oryx & - & - & - & - & - & - & 15.0 & 37.3 & - & - & - & - \\
LL3DA & - & - & - & - & 36.8 & 65.2 & 15.9 & 37.3 & - & - & - & - \\
LEO & - & - & - & - & \underline{38.2} & 72.4 & \textcolor{lightgray}{20.0} & \textcolor{lightgray}{49.2} & - & - & 50.0 & 52.4 \\
Scene-LLM & - & - & - & - & - & - & 16.6 & 40.0 & - & - & 54.2 & - \\
Chat-3D v2 & 35.9 & 30.4 & - & - & 15.5 & 28.1 & 16.1 & 40.1 & - & - & - & - \\
Grounded-3DLLM & 47.9 & 44.1 & 45.2 & 40.6 & 35.5 & 70.6 & 15.2 & 37.1 & - & - & - & - \\
Chat-Scene & \underline{55.5} & 50.2 & \underline{57.1} & \underline{52.4} & 36.3 & 77.1 & \underline{18.0} & \underline{41.6} & \underline{53.2} & \underline{56.1} & \underline{54.6} & \underline{57.5} \\
\rowcolor{green!15} Robin3D (Ours) & \textbf{60.8} & \textbf{55.1} & \textbf{64.9} & \textbf{59.7} & \textbf{38.4} & \textbf{87.2} & \textbf{19.2} & \textbf{44.0} & \textbf{56.0} & \textbf{58.6} & \textbf{56.9} & \textbf{59.8} \\ \bottomrule
\end{tabular}}
\caption{\textbf{Quantitative comparison.} 
``Task-Specific Training'' denotes models trained on a specific task, while ``Joint Training'' denotes models trained jointly on multiple tasks.
Entries in \textcolor{lightgray}{gray} denote using ground truth question-relative objects annotations. The best and second best results in a fair comparison are highlighted in \textbf{bold} and \underline{underline}, respectively.}
\label{final performance}
\end{table*}

\begin{figure}[t] %
    \centering
    \includegraphics[width=0.48\textwidth]{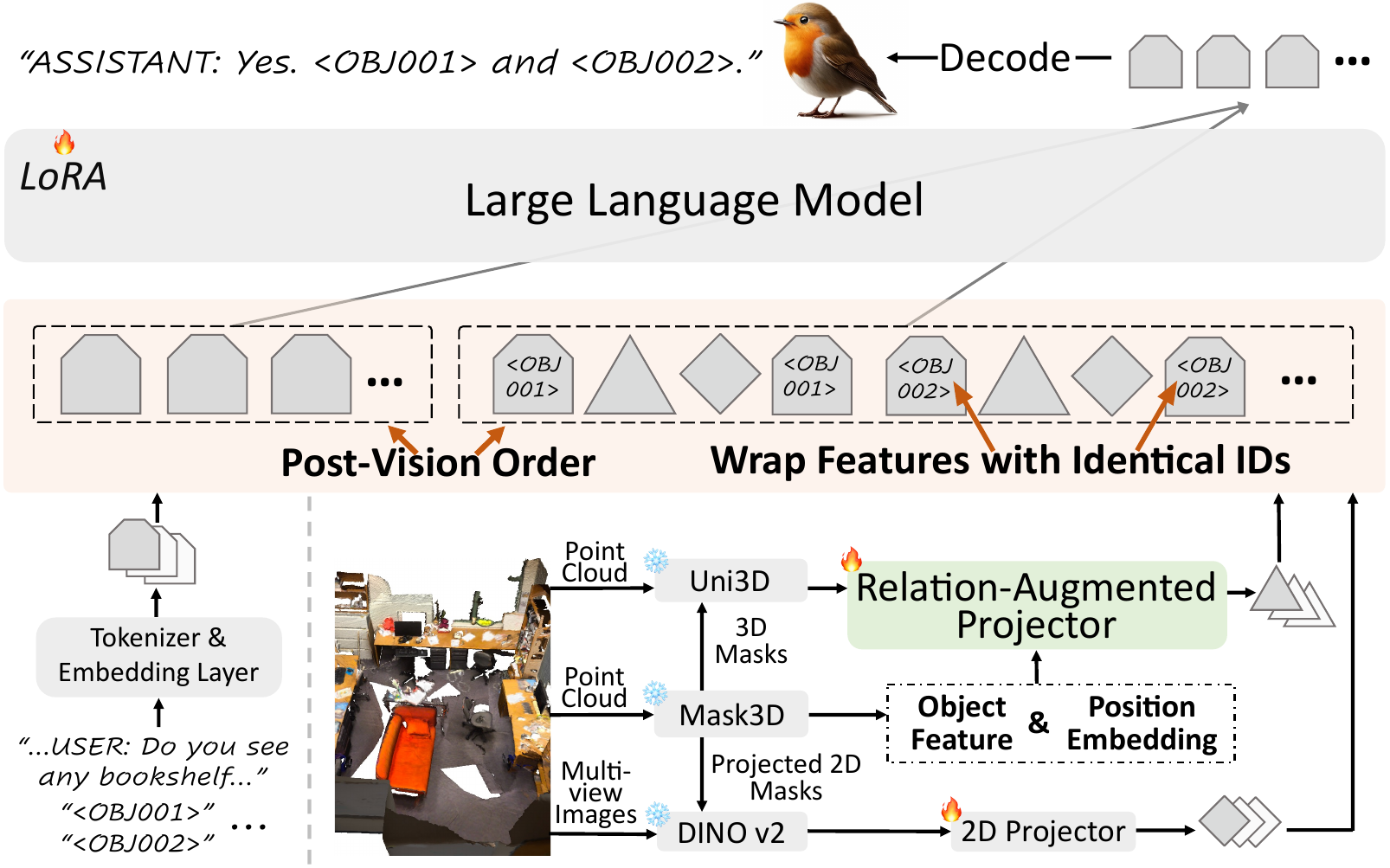}
    \vspace{-15pt}
    \caption{Overview of Robin3D model structure. 
    \textit{Bottom}: \colorbox[rgb]{0.886, 0.941, 0.851}{Relation-Augmented Projector} fuses the features and position embedding from Mask3D and Uni3D for final 3D features. 2D features from DINO v2 are projected into the LLM space.
    \textit{Middle}: \colorbox[rgb]{0.984, 0.898, 0.839}{Sequence Organization} enhances the connection between object IDs and object features by wrapping the features with identical IDs and the Post-Vision order.
    \textit{Top}: We use LoRA to fine-tune the LLM on our constructed 1 million instruction data.}
\label{method}
\vspace{-10pt}
\end{figure}

\subsection{Improvements on the model structure}
We adopt Chat-Scene \citep{chat-scene} as the baseline of our model structure, and further provide improvements on the vision projector and the organization of the sequence of tokens. Detailed and formal demonstration of our model structure is put in the Supplementary to avoid redundancy. We illustrate our improvements in the followings.

\paragraph{Vision Projector}
Chat-Scene's \citep{chat-scene} 3D features inevitably weaken the spatial relationships between objects due to the object-centric normalization in Uni3D \citep{uni3d}, which hinders the learning on our diverse visual grounding data.
As shown in Fig.~\ref{method} (bottom), to obtain relation-aware 3D features while preserving the unified object-centric characteristics, our Relation-Augmented
Projector (RAP) considers three types of 3D features: a) the object features from Mask3D, $X_{mask3d}$, which are scene-level respresentations, containing spatial relationship, as they come across multiple cross-attention layers to exchange information. b) the position embeddings of the Mask3D, $X_{pos}$, which are directly projected from the objects' coordinates. c) the unified object features, $X_{uni3d}$, from Uni3D. Finally, our RAP is formulated as:
\begin{equation}
\begin{split}
    X = \text{Concat}(&\text{Norm}_{L2}(X_{uni3d}), \text{Norm}_{L2}(X_{mask3d})), \\
    X_{rap} &= \text{MLP}(X) + \text{MLP}(X_{pos})
\end{split}
\end{equation}
where $\text{Norm}_{L2}$ is the L2 normalization, $\text{Concat}$ is the concatenation alongside the channel dimension, and $\text{MLP}$ is a multi-layer perceptron with GELU activation \citep{gelu}. $X_{rap}$ represents our final 3D features.

\paragraph{Sequence Organization}
Previous approaches \citep{chat3dv2, chat-scene} simply append object ID to object feature as a prefix, which may lack sufficient connection between the ID and the feature, making it struggle with complex referring and grounding requirements in our adversarial instruction data.
Here, we provide a more informative way to organize the input sequence.
As shown in Fig.~\ref{method} (middle), we first use two identical ID tokens to wrap the object features. Adhering to the causal attention nature of the LLM, this approach links ID information to the object features via the first ID token, and links object information to its ID via the second ID token. 
Secondly, we adopt a post-vision order, which places the vision tokens at the end of the input sequence, closer to the answer tokens generated by the model. This approach mitigates attention deviation from the answer tokens to the ID-Feature pairs, a problem caused by their relative token distance and rotary position embeddings \citep{rope, vista}.
The post-vision order is structured as:
[\textit{$<$Question tokens$>$, $<$Vision tokens$>$, $<$Answer tokens$>$}], where $<$Vision tokens$>$ comprises the ID tokens and object feature tokens.
\section{Experiments}
In this section, we present the main quantitative results and ablation studies. We have also conducted a comprehensive analysis of our Robust Instruction Data, along with more detailed ablation studies. However, due to space limitations, these are provided in the Supplementary. We highly recommend that readers refer to it for a complete understanding.

\begin{figure}[t]
    \centering
    \includegraphics[width=0.48\textwidth]{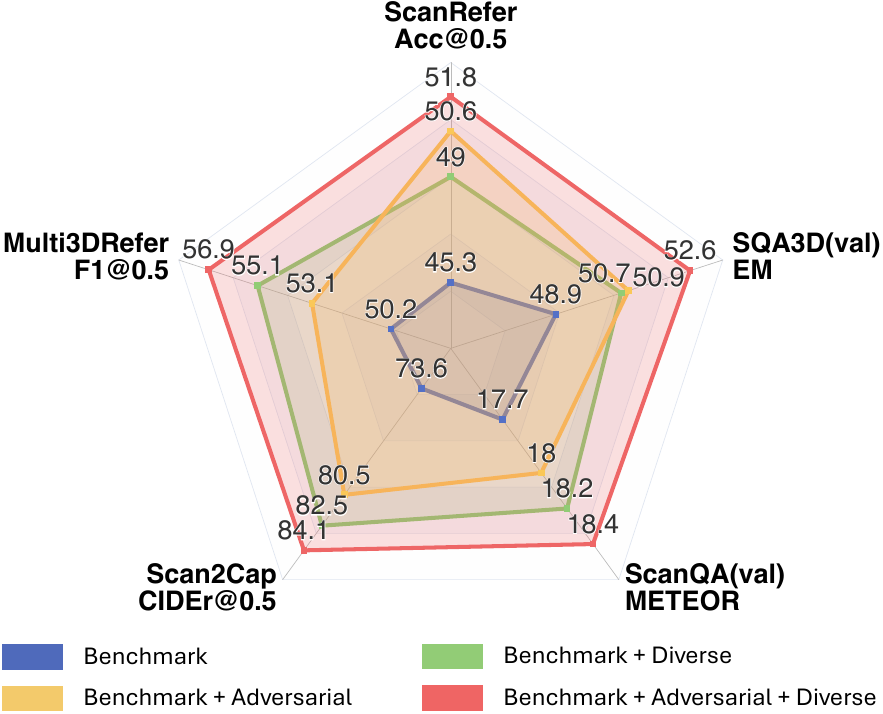}
    \vspace{-14pt}
    \caption{\textbf{Ablation study on Robust Instruction Generation.} \textit{Benchmark} denotes training on the original training set of the benchmarks. \textit{Adversarial} denotes adding the Adversarial Instruction data to the training set. \textit{Diverse} denotes adding the Diverse Instruction data to the training set.}
\label{ablation_rig}
\vspace{-8pt}
\end{figure}

\begin{figure}[t]
    \centering
    \includegraphics[width=0.48\textwidth]{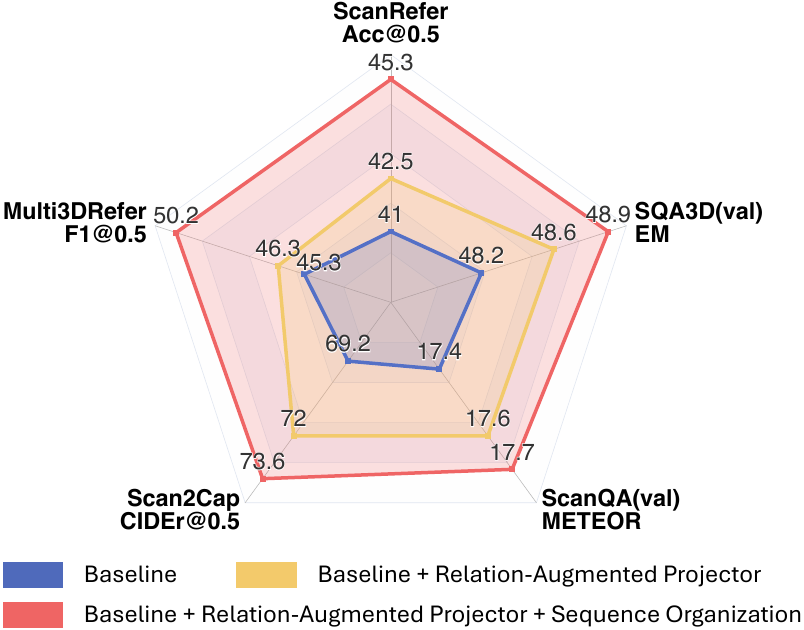}
    \vspace{-14pt}
    \caption{\textbf{Ablation study on our proposed improvements:} Relation-Augmented Projector and Sequence Organization.}
    \label{ablation_method}
    \vspace{-10pt}
\end{figure}

\subsection{Benchmarks and Metrics}
We provide quantitative results on the widely-used 3D multimodal learning datasets: ScanRefer \citep{scanrefer} for 3D Visual Grounding, Multi3DRefer \citep{multi3drefer} for General 3D Visual Grounding including zero, single and multiple target objects, Scan2Cap \citep{scan2cap} for 3D Dense Captioning, ScanQA \citep{scanqa} for 3D Question Answering, and SQA3D \citep{sqa3d} for 3D Situated Question Answering. The vision data are all based on the ScanNet dataset \citep{scannet}, which contains real world 3D point clouds across 1,513 indoor scenes with detailed object annotations. All these benchmarks follow the same data split as ScanNet.

We follow the standard evaluation metrics widely adopted in the respective benchmarks. For ScanRefer, we calculate accuracy at Intersection over Union (IoU) thresholds of 0.25 and 0.5 (Acc@0.25, Acc@0.5). For Multi3DRefer, we use the F1 score with IoU thresholds of 0.25 and 0.5 to measure performance. In Scan2Cap, we apply the CIDEr@0.5 and BLEU-4@0.5 (C@0.5, B-4@0.5) metrics, combining standard captioning metrics with the IoU metric. For ScanQA, the METEOR and ROUGE metrics, denoted as M and R, are employed. Lastly, SQA3D is assessed with exact match accuracy (EM) and its extended form, EM-R, as suggested by LEO \citep{leo}.

\subsection{Implementation Details}
We extract 150 object features from each 3D scene, along with the corresponding position embeddings and 3D masks generated by Mask3D. 
Following Chat-Scene's setup, we use a two-layer MLP as the 2D Projector and the Vicuna-7B-v1.5 model \citep{vicuna} as our LLM. We fine-tune the model using LoRA \citep{lora} (with a rank of 16) by Cross Entropy loss.
The global learning rate is formulated as [batch size $\times$ base learning rate $\times$ number of GPUs] and is set to 0.00064, with a cosine annealing schedule. 
For our results in Tab.~\ref{final performance}, we train 2 epochs on the RIG-generated data, and then train 2 epochs on the benchmark data in the second stage.
We train 1 epoch for each stage to efficiently conduct ablation studies of RIG-generated data.
For ablation on our improvements of the model structure, we train 1 epoch on the benchmark data to avoid potential compound effects.

\begin{figure*}[t]
    \centering
    \includegraphics[width=1\textwidth]{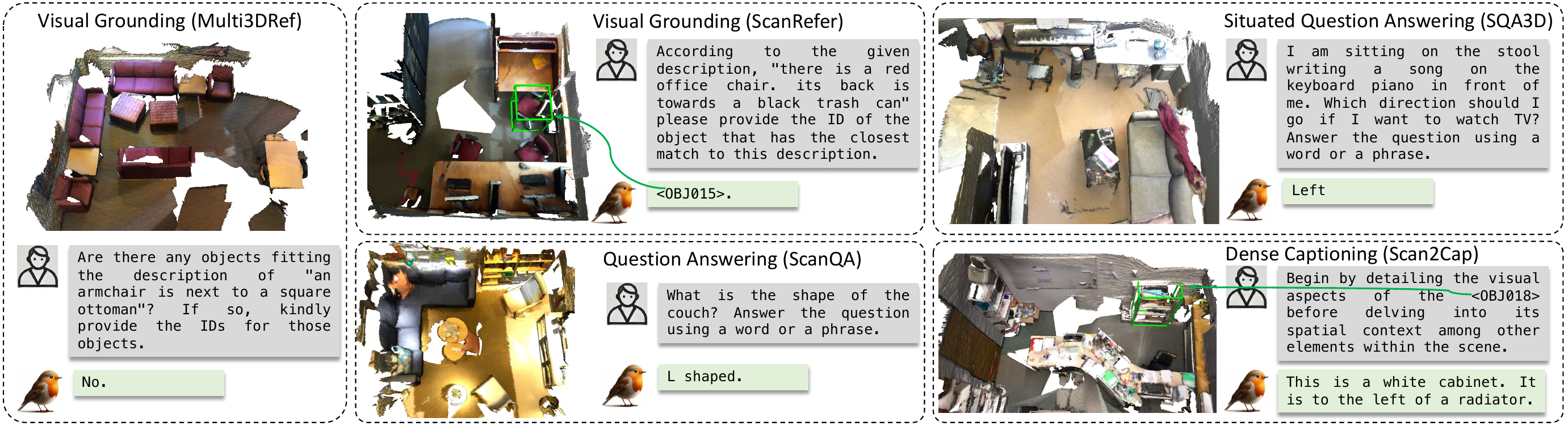}
    \vspace{-20pt}
    \caption{Visualization of Robin3D's responses on all the benchmarks.}
    \label{visualization}
    \vspace{-15pt}
\end{figure*}

\subsection{Quantitative Results}
We classify current methods into two categories: Task-Specific Training and Joint Training. Task-Specific Training refers to models only trained for a specific task, while Joint Training means training on multiple tasks jointly. Our Robin3D does not conduct task-specific fine-tuning.

\begin{itemize}[noitemsep, leftmargin=*]
    \item \textbf{Task-Specific Training:} As shown in Table~\ref{final performance}, models like EDA and M3DRef-CLIP perform well on their respective tasks due to customized model design for the task. However, they lack the ability to generalize to other tasks. Models like Vote2Cap-DETR++ and SQA3D encounter the similar issue. Therefore, they are not suitable to serve as general-purpose 3D AI agents.
    \item \textbf{Joint Training:} Benefiting from sharing the knowledge across multiple tasks, models like 3D-VisTA and PQ3D show decent performance across multiple tasks, but their dependence on task-specific heads restricts their generalizability. Models like LEO and Chat-Scene show promising results by leveraging LLMs, but their sole training on positive pairs and template-based instructions leads to suboptimal generalization.
    \item \textbf{Our Robin3D:} Due to the robust instruction data generated by RIG, Robin3D significantly outperforms previous models across all the benchmarks. Specifically, Robin3D brings a \textbf{6.9\%} improvement on Scan2Cap CIDEr@0.5 and a \textbf{5.3\%} improvement on ScanRefer Acc@0.25. Notably, on the evaluation of Multi3DRefer, which contains zero-target cases that are challenging for models to be discriminative and learn to say ``{\em No}'', our Robin3D achieves a \textbf{7.8\%} improvement in F1@0.25 and a \textbf{7.3\%} improvement in F1@0.5.
\end{itemize}

\subsection{Ablation Study}

We perform ablation studies on the training data and model structure, respectively. We first evaluate the effectiveness of RIG-generated data by progressively adding the Adversarial Instruction data and the Diverse Instruction data to the training set. We then investigate the contribution of Relation-Augmented Projector and Sequence Organization by comparing models with and without these components.

\noindent\textbf{Robust Instruction Generation (RIG):} As shown in \cref{ablation_rig}, by adding the Adversarial Instruction data, we observe a consistent improvement across all benchmarks. Specifically, performance on ScanRefer and Multi3DRefer increases by \textbf{3.7\%} and \textbf{4.9\%}, respectively. It is worth noting that the performance on Scan2Cap improves by \textbf{8.9\%}, even though there is not any object captioning data in the Adversarial Instruction data, which highlights its contribution on enhancing the understanding towards each object by challenging the model with mixed positive and negative samples. 
Additionally, by adding the Diverse Instruction data, we also provide comprehensive improvements. Specifically, descriptions from the original ScanRefer are annotated by human following a fixed instruction template or expression style, which limits the language diversity. In contrast, the Diverse Instruction data contains various language styles and task formats, which helps the model generalize better, resulting in a \textbf{5.3\%} improvement on ScanRefer. Finally, by combining both two types of data, we achieve a further improvement, demonstrating the effectiveness of RIG-generated data.

\noindent\textbf{Relation-Augmented Projector and Sequence Organization:} \label{model_structure}
As shown in \cref{ablation_method}, the baseline indicates the Chat-Scene's structure and the integration of our projector leads to steady improvements across all benchmarks. Notably, the performance on Visual Grounding tasks, including ScanRefer and Multi3DRefer, shows significant gains due to our enhanced spatial comprehension. When adopting our sequence organization, further improvements are observed, emphasizing the importance of refining the model's object referring and grounding capabilities by reinforcing the connection between IDs and features.

\subsection{Qualitative Results}
We provide the visualization of Robin3D's responses on all the benchmarks in Fig.~\ref{visualization} with the prompts of each task. These results demonstrate the generalization ability of Robin3D on various tasks.

\section{Conclusion}
\vspace{-3pt}
We identify the problem of a lack of robust instruction training data in current 3DLLMs. To tackle this challenge, we introduce Robin3D, a powerful 3DLLM trained on large-scale instruction-following data generated by our novel data engine, Robust Instruction Generation (RIG) engine. We generate and collect 1 million instruction data, including benchmark data, adversarial data, and diverse data. 
To better handle our complex data, Robin3D incorporates a Relation-Augmented Projector and enhanced sequence organization
for better object referring and grounding. Finally, Robin3D achieves state-of-the-art performance across all the widely-used 3D multimodal learning benchmarks.

\section{Future Work} 
\vspace{-3pt}
Our main contribution lies in the novel approach of constructing robust data to enhance the model's capabilities. We hope this work will inspire the current community to explore ways to further improve the complexity and quality of training data, rather than simply relying on available benchmark datasets. At the same time, developing a rigorous and comprehensive benchmark to better evaluate model performance is also a potentially valuable direction.

Additionally, our study follows previous common setting in 3DLLM \cite{3dllm, ll3da, chat3dv2, chat-scene} by training and testing on the ScanNet indoor environment. Expanding future work to outdoor environments or to data based on LiDAR could further close the gap in 3DLLM and Spatial Intelligence.

It is worth noting that open-vocabulary capabilities have not yet been formally evaluated or emphasized in previous 3DLLM research. Training a powerful vision backbone that simultaneously supports open-vocabulary capabilities, unified representation, and scene-level representation could foreseeably lead to significant improvements.

\clearpage
{
    \small
    \bibliographystyle{ieeenat_fullname}
    \bibliography{main}
}


\clearpage
\setcounter{page}{1}
\twocolumn[{%
\renewcommand\twocolumn[1][]{#1}%
\maketitlesupplementary
\begin{center}
    \centering
    \captionsetup{type=figure}
    \includegraphics[width=0.93\textwidth]{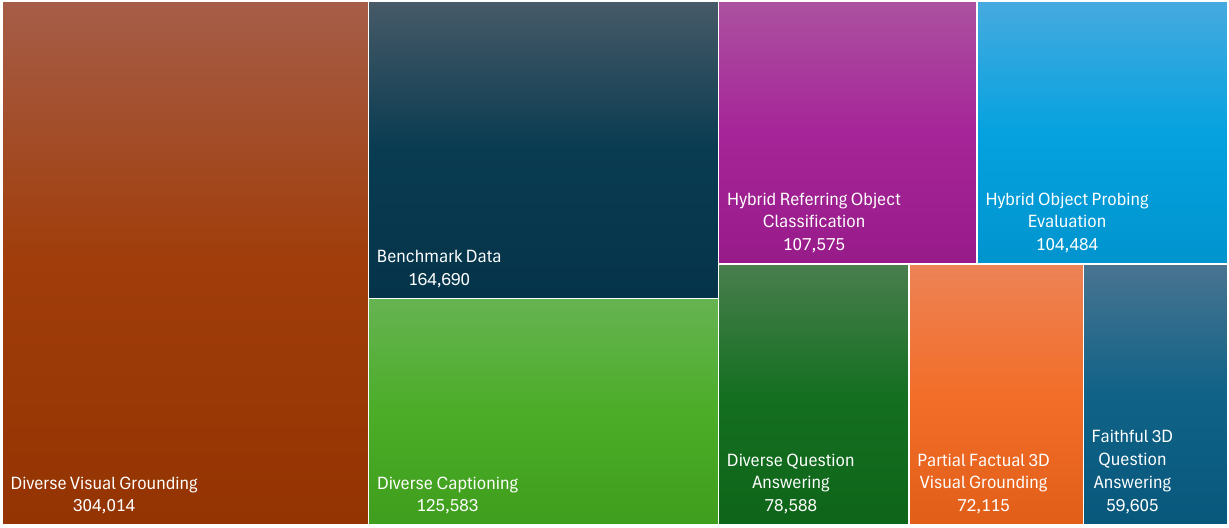}
    \vspace{-10pt}
    \captionof{figure}{The number of samples for different tasks in our robust data and the visualization of their proportion of the total data.}
    \label{distribution}
\end{center}
}]

\begin{figure}[t]
    \centering
    \includegraphics[width=0.48\textwidth]{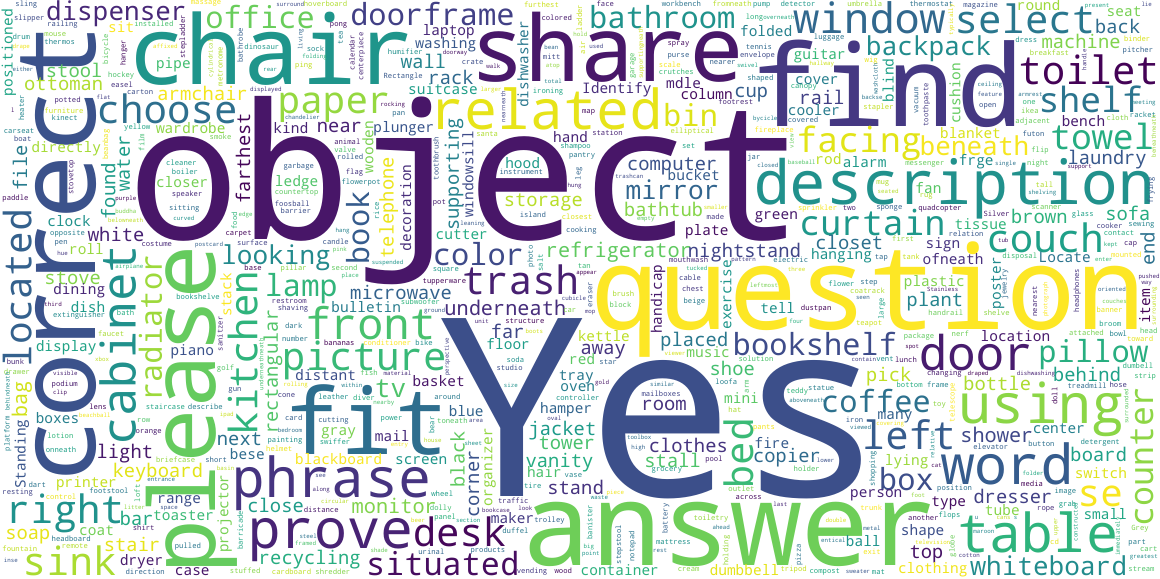}
    \vspace{-18pt}
    \caption{The word cloud of our adversarial data.}
    \label{adversarial_wordcloud}
    \vspace{-8pt}
\end{figure}

\begin{figure}[t]
    \centering
    \includegraphics[width=0.48\textwidth]{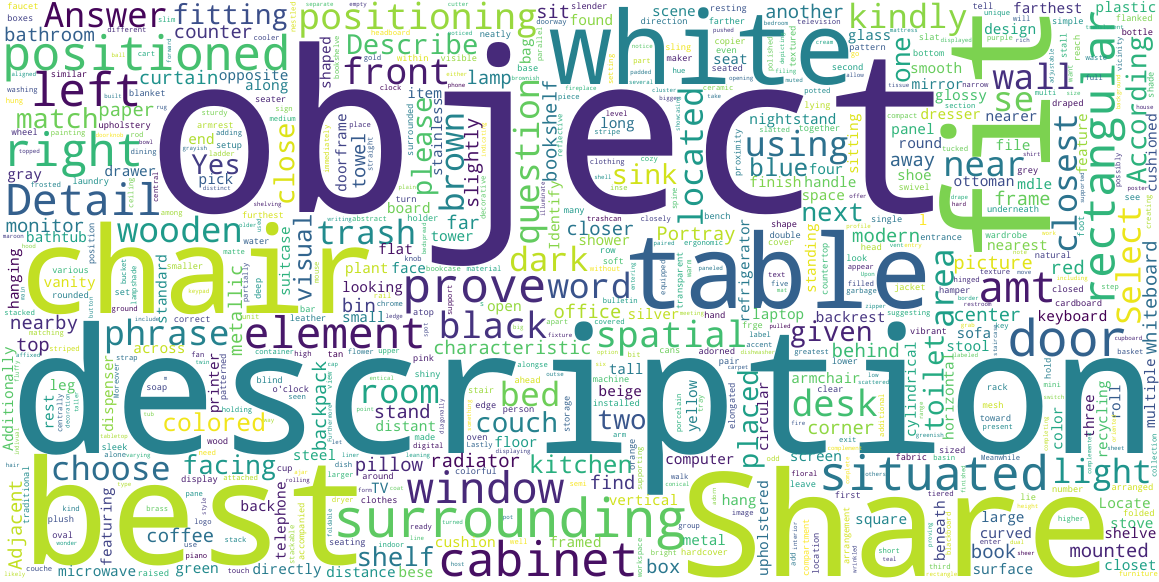}
    \vspace{-18pt}
    \caption{The word cloud of our diverse data.}
    \label{diverse_wordcloud}
    \vspace{-14pt}
\end{figure}

\begin{figure*}[t]
    \centering
    \includegraphics[width=0.88\textwidth]{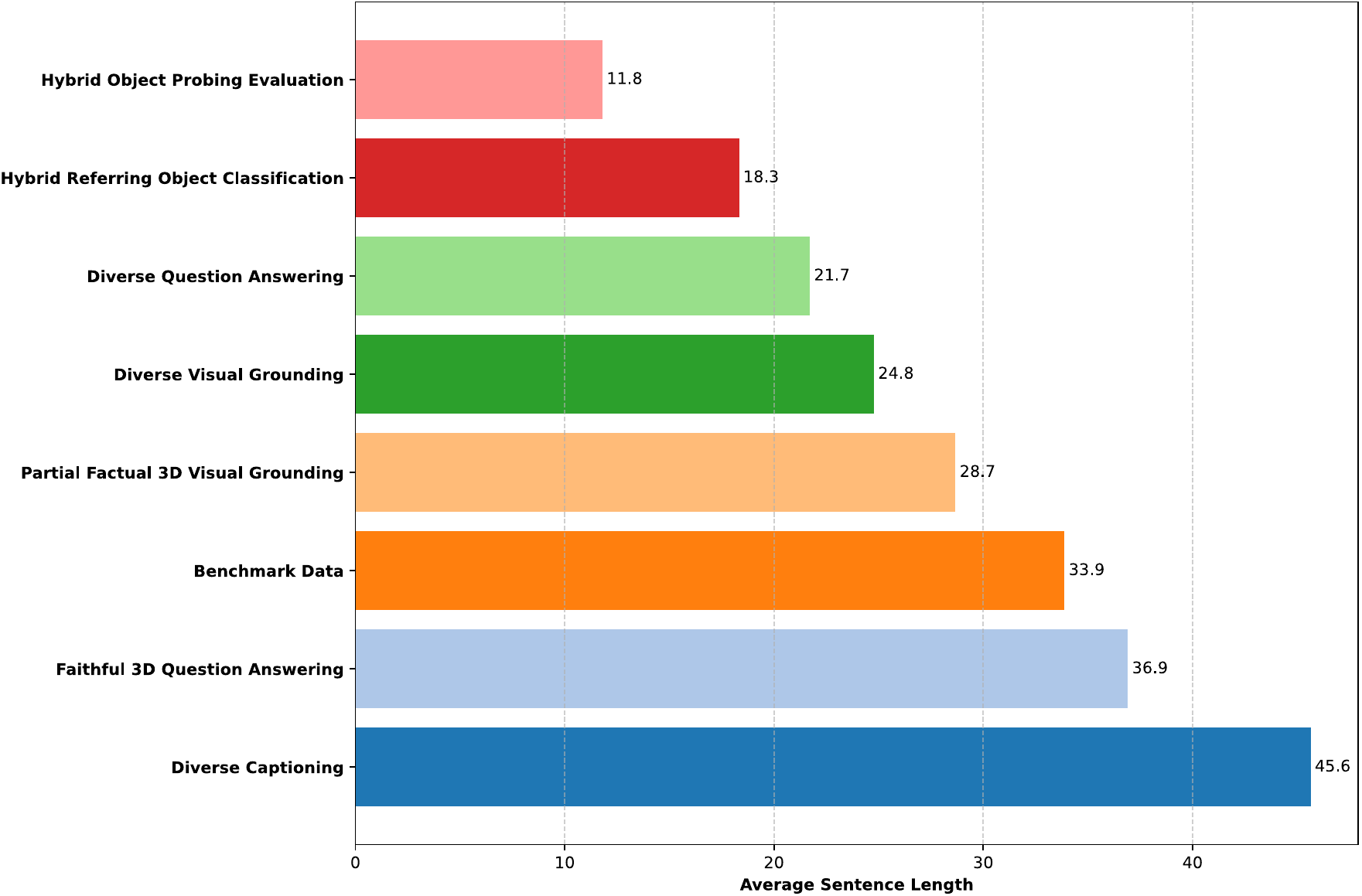}
    \caption{The average sentence length of different tasks in our robust data.}
    \label{length}
\end{figure*}

\section{Data Analysis}
As shown in \cref{distribution}, we provide detailed statistics on the number of samples for different tasks in our robust dataset, along with a qualitative result of their respective proportions in the total data. For our Diverse Instruction data, we split it into three parts based on the task categories for statistical purposes, which include Diverse Visual Grounding, Diverse Captioning, and Diverse Question Answering.

We further present the word cloud of our Adversarial Instruction data and Diverse Instruction data in \cref{adversarial_wordcloud} and \cref{diverse_wordcloud}, respectively. We exclude the words related to object IDs, as they pertain to the referring and grounding format rather than the actual data content.

In \cref{length}, we provide statistics on the average sentence length for each task in our robust dataset. Here, the sentence length is calculated as the number of words in the question prompt plus the number of words in the answer, excluding the count of object IDs.

\section{Detailed Ablation Study of Adversarial data}
To further evaluate the effectiveness of each task in our Adversarial Instruction data, we conduct detailed ablation studies on Hybrid Object Probing Evaluation data, Hybrid Referring Object Classification data, Partial Factual 3D Visual Grounding data, and Faithful 3D Question Answering data by adding them to the benchmark data in each experiment. As shown in \cref{rig}, all four tasks in the Adversarial data contribute notable improvements compared with solely training on the benchmark data.

\begin{table*}[h]
    \centering
    \resizebox{0.9\linewidth}{!}{
    \begin{tabular}{cccccc}
    \toprule
    \multirow{2}{*}{\textbf{Data}} & ScanRefer & Multi3DRefer & Scan2Cap & ScanQA(val) & SQA3D(val) \\
     & Acc@0.5 & F1@0.5 & C@0.5 & M & EM \\ \midrule
    Benchmark & 45.3 & 50.2 & 73.6 & 17.7 & 48.9 \\
    \rowcolor{green!5} + HOPE & 45.8 & 52.5 & 76.1 & 17.8 & 50.1 \\
    \rowcolor{green!5} + HROC & 47.7 & 53.0 & 78.9 & 18.0 & 50.3 \\
    \rowcolor{green!5} + PF-3DVG & 45.7 & 51.0 & 77.2 & 17.9 & 49.6 \\
    \rowcolor{green!5} + 3DFQA & 47.2 & 52.1 & 77.5 & 17.9 & 50.2 \\ \bottomrule
    \end{tabular}}
    \caption{\textbf{Ablation study on Adversarial Instruction data.} \textit{Benchmark} denotes training on the original training set of the benchmarks. \textit{HOPE} denotes adding the Hybrid Object Probing Evaluation data to the original training set. \textit{HROC} denotes adding the Hybrid Referring Object Classification data to the original training set. \textit{PF-3DVG} denotes adding the Partial Factual 3D Visual Grounding data to the original training set. \textit{3DFQA} denotes adding the Faithful 3D Question Answering data to the original training set.}
    \label{rig}
\end{table*}

\begin{figure*}
    \centering
    \captionsetup{type=figure}
    \includegraphics[width=.93\textwidth]{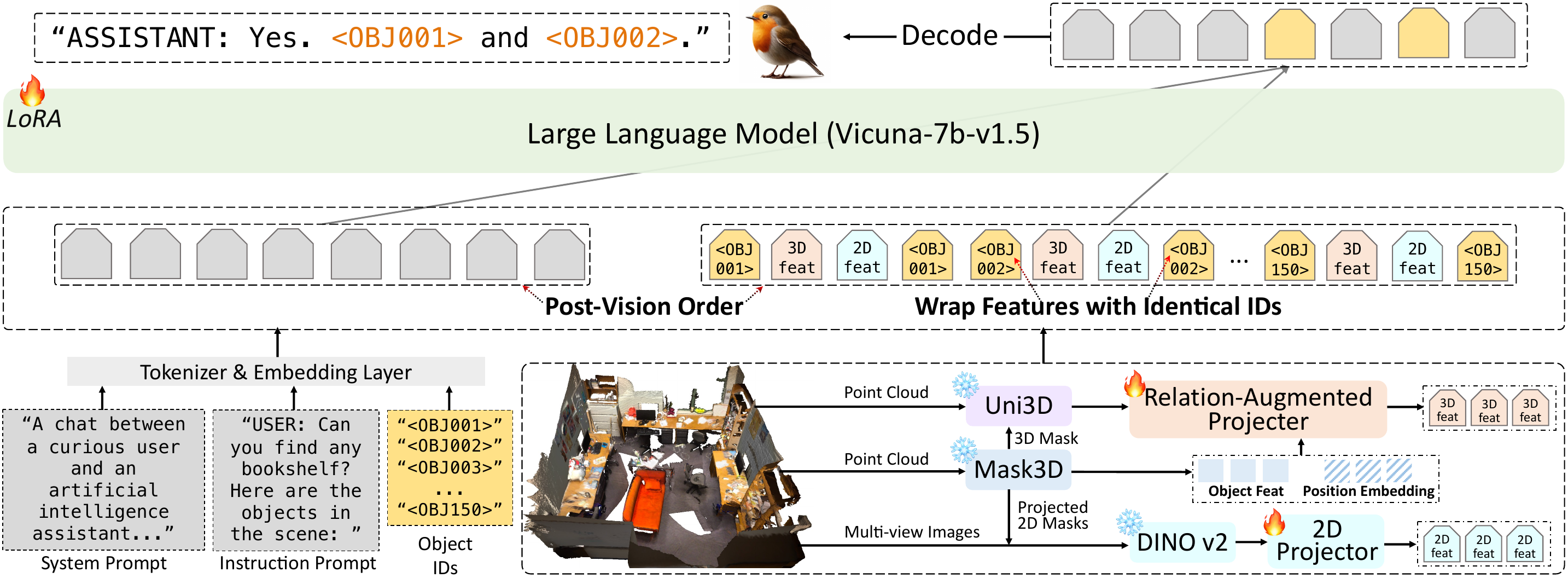}
    \captionof{figure}{Overview of Robin3D model structure. 
    \textit{Bottom}: Our Relation-Augmented Projecter fuses the features and position embedding from Mask3D and Uni3D to generate final 3D features. 2D features from DINO v2 are projected into the LLM space. We freeze the Mask3D, Uni3D, and DINO v2.
    \textit{Middle}: We enhance the connection between object IDs and object features by wrapping the features with identical IDs and the Post-Vision order.
    \textit{Top}: We use LoRA to fine-tune the LLM on our constructed 1 million instruction data.}
    \label{model_detail}
\end{figure*}%

\section{Details of Robin3D}
To train a 3D LLM using instruction fine-tuning, we first represent the 3D scene as a sequence of vision tokens, then append it with system and instruction prompts, expressed as sequences of language tokens, to indicate the task. Taking the above tokens as input, a LLM is supervised to output the answer tokens via next token prediction.
Specifically,
as shown in \cref{model_detail}, given the point cloud of a 3D scene, we use the pre-trained 3D segmenter Mask3D \citep{mask3d} to extract object features along with their corresponding 3D masks. Following Chat-Scene \citep{chat-scene}, we further sample each object's point cloud based on the 3D masks, normalize it, and employ the pre-trained Uni3D \citep{uni3d} to extract unified object-centric 3D features. Additionally, 2D masks projected from the 3D masks are used to sample and average 2D features, which are extracted by DINO v2 from multi-view images of each object. 
Our Relation-Augmented Projector fuses the 3D features and position embeddings from Mask3D and Uni3D into our final 3D features. 
In line with Chat-Scene \citep{chat-scene}, we incorporate special tokens $\{<\texttt{OBJ}_i>\}_{i=1...n}$ as object IDs into the vocabulary. These ID tokens are paired with 2D and 3D object features to indicate each object, for referring to the object in the input instruction or grounding the object in model's output.
We combine each object feature with its corresponding object ID, and appends the system and question prompts at the beginning of the sequence, which are then fed into the LLM.

\end{document}